\documentclass[10pt,twocolumn,letterpaper]{article}

\usepackage{cvpr}
\usepackage{times}
\usepackage{epsfig}
\usepackage{graphicx}
\usepackage{amsmath}
\usepackage{amssymb}
\usepackage{balance}

\begin{document}

\title{iWarpGAN: Disentangling Identity and Style to Generate Synthetic Iris Images}

\author{Shivangi Yadav\\
Michigan State University\\
{\tt\small yadavshi@msu.edu}
\and
Arun Ross\\
Michigan State University\\
{\tt\small rossarun@msu.edu}
}

\maketitle
\thispagestyle{empty}

\begin{abstract}
   Generative Adversarial Networks (GANs) have shown success in approximating complex distributions for synthetic image generation. However, current GAN-based methods for generating biometric images, such as iris, have certain limitations: (a) the synthetic images often closely resemble images in the training dataset; (b) the generated images lack diversity in terms of the number of unique identities represented in them; and (c) it is difficult to generate multiple images pertaining to the same identity. To overcome these issues, we propose iWarpGAN that disentangles identity and style in the context of the iris modality by using two transformation pathways: Identity Transformation Pathway to generate unique identities from the training set, and Style Transformation Pathway to extract the style code from a reference image and output an iris image using this style. By concatenating the transformed identity code and reference style code, iWarpGAN generates iris images with both inter- and intra-class variations. The efficacy of the proposed method in generating such iris DeepFakes is evaluated both qualitatively and quantitatively using ISO/IEC 29794-6 Standard Quality Metrics and the VeriEye iris matcher. Further, the utility of the synthetically generated images is demonstrated by improving the performance of deep learning based iris matchers that augment synthetic data with real data during the training process.
\end{abstract}

\vspace{-5mm}
\section{Introduction}
\label{sec:intro}
The iris is a thin, circular structure in the eye that controls the amount of light that enters the eye by adjusting the size of the pupil. It is located in front of the lens and behind the cornea and is composed of muscles and pigmented tissue. The distinctive nature of the iris pattern has led to its use as a reliable biometric cue in identification and authentication systems \cite{nigam2015}. With the advent of technology, iris sensors are now available in commercial and personal devices, paving the way for secure authentication and access control \cite{gent2023, irisgalaxy8}. However, the accuracy of iris recognition systems relies heavily on the quality and size of the dataset used for training. The limited availability of large-scale iris datasets due to the difficulty in collecting operational quality iris images, has become a major challenge in this field. For example, most of the iris datasets available in the literature have frontal view images \cite{casia1000, kumar2010}, and the number of subjects and total number of samples in these datasets are limited. Further, in some instances, collecting and sharing iris datasets may be stymied due to privacy or legal concerns \cite{voigt2017}.  Therefore, researchers have been studying the texture and morphology of the iris in order to model its unique patterns and to create large-scale {\em synthetic} iris datasets. For example, Cui et al. \cite{cui2004} utilized principal component analysis with super-resolution to generate synthetic iris images. Shah and Ross \cite{shah2006} used a Markov model to capture and synthesize the iris texture followed by embedding of elements such as spots and stripes to improve visual realism. In \cite{zuo2007}, Zuo et al. analyzed various features of real iris images, such as texture, boundary regions, eyelashes, etc. and used these features to create a generative model based on the Hidden Markov Model for synthetic iris image generation. These methods while successfully generating synthetic iris images, are found lacking in terms of quality (visual realism and good-resolution) and diversity in the generated samples \cite{yadav2019}.

\begin{figure}
    \centering
    \setlength\belowcaptionskip{0pt}
    \captionsetup{justification=centering}
    \includegraphics[width=0.45\textwidth]{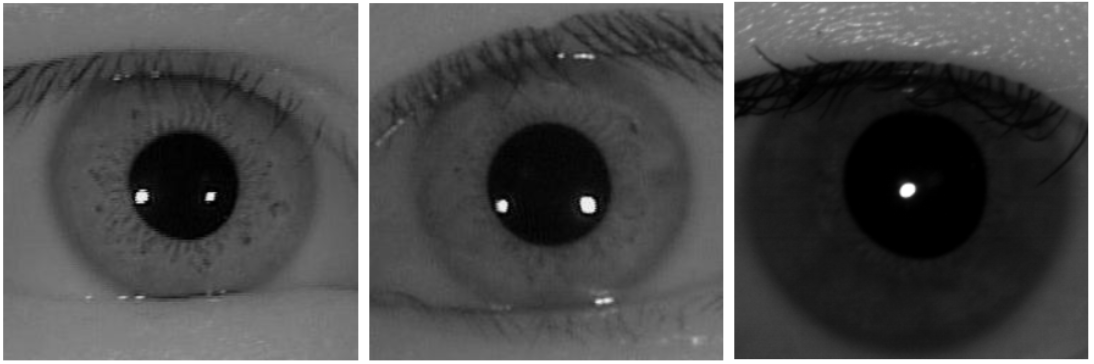}
    \caption{Examples of real cropped iris images from publicly available datasets \cite{casia1000}\cite{iitd-iris}\cite{xiao2013}.}
    \vspace*{-5mm}
    \label{fig:mesh1}
\end{figure}


Over the past few years, deep learning-based approaches have set a benchmark in various fields including synthetic image generation and attribute editing, using Convolutional Autoencoders (CAEs) \cite{van2016} and Generative Adversarial Networks (GANs) \cite{goodfellow2020,lee2021}. In \cite{kohli2017,yadav2019,yadav2021}, authors proposed GAN-based synthetic image generation methods that input a random noise vector and output a synthetic iris image. While these methods address some of the concerns mentioned previously, the generated images are often similar to each other \cite{patrick2022}. Additionally, due to insufficient number of training samples, the generator is often over-trained to synthesize images with patterns seen during training \cite{patrick2022}, which affects the uniqueness of the synthesized iris images (as shown in Figure \ref{fig:Exp1}).

In this paper, we address the following  limitations of current synthetic iris generators: (1) difficulty in generating good quality synthetic iris images, (2) failure to incorporate inter and intra class variations in the generated images, (3) generating images that are similar to the training data, and (4) utilizing domain-knowledge to guide the synthetic generation process. We achieve this by proposing iWarpGAN that aims to disentangle identity and style using two transformation pathways: (1) Identity Transformation and (2) Style Transformation. The goal of Identity Transformation pathway is to transform the identity of the input iris image in the latent space to generate identities that are different from the training set. This is achieved by learning RBF-based warp function, \ensuremath{f^p}, in the latent space of a GAN, whose gradient gives non-linear paths along the \ensuremath{p^{th}} family of paths for each latent code \ensuremath{z \in \mathbb{R^d}}. The Style Transformation pathway aims to generate images with different styles, which are extracted from a reference iris image, without changing the identity. Therefore, by concatenating the reference style code with the transformed identity code, iWarpGAN generates iris images with both inter and intra-class variations. Thus, the contributions of this research are as follows:

\noindent
(a) We propose a synthetic image generation method, iWarpGAN, which aims to disentangle identity and style in two steps: identity transformation and style transformation. \\
\noindent
(b) We evaluate the quality and realism of the generated iris images using ISO/IEC 29794-6 Standard Quality Metrics which uses a non-reference single image quality evaluation method. \\
\noindent
(c) We show the utility of the generated iris dataset in training deep-learning based iris matchers by increasing the number of identities and overall images in the dataset.
\vspace{1mm}

In the remainder of this paper, we will discuss the proposed method in more detail and demonstrate the advantages of the proposed method in generating good quality, unique iris images compared to other GAN-based methods.


\begin{figure*}
    \centering
    \captionsetup{justification=centering}
    \includegraphics[width=0.75\textwidth]{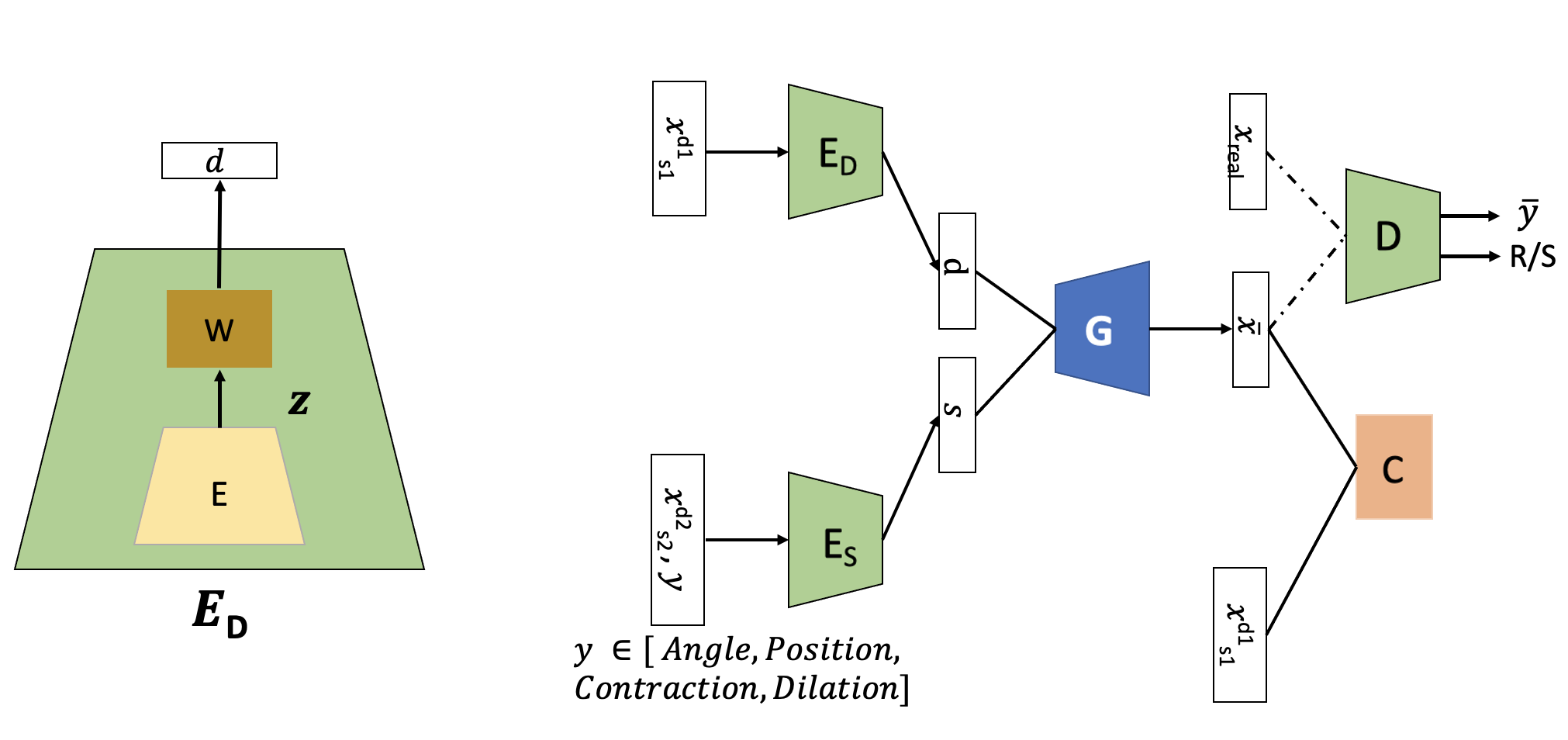}
    \caption{The proposed iWarpGAN consists of five parts: (1) Style Encoder, \ensuremath{E_S}, that aims to encode the style of the input image as \ensuremath{s}, (2) Identity Encoder, \ensuremath{E_D}, that aims to learn encoding \ensuremath{d} that generates an identity different from the input image, (3) Generative Network, \ensuremath{G}, that uses encoding from both \ensuremath{E_D} and \ensuremath{E_S} to generate an image with a unique identity and the given style attribute, (4) Discriminator, \ensuremath{D}, that inputs either a real or synthetic image and predicts whether the image is real or synthetic and also emits an attribute vector \ensuremath{y^{'} \in} \{angle, position, contraction, dilation of pupil\}, and (5) Pre-trained Classifier, \ensuremath{C}, that computes the distance score between the real input image and the new identity generated by \ensuremath{G}.}
    \vspace{-5mm}
    \label{fig:iWarpGAN}
\end{figure*}

\begin{figure}
    \centering
    \captionsetup{justification=centering}
    \includegraphics[width=0.47\textwidth]{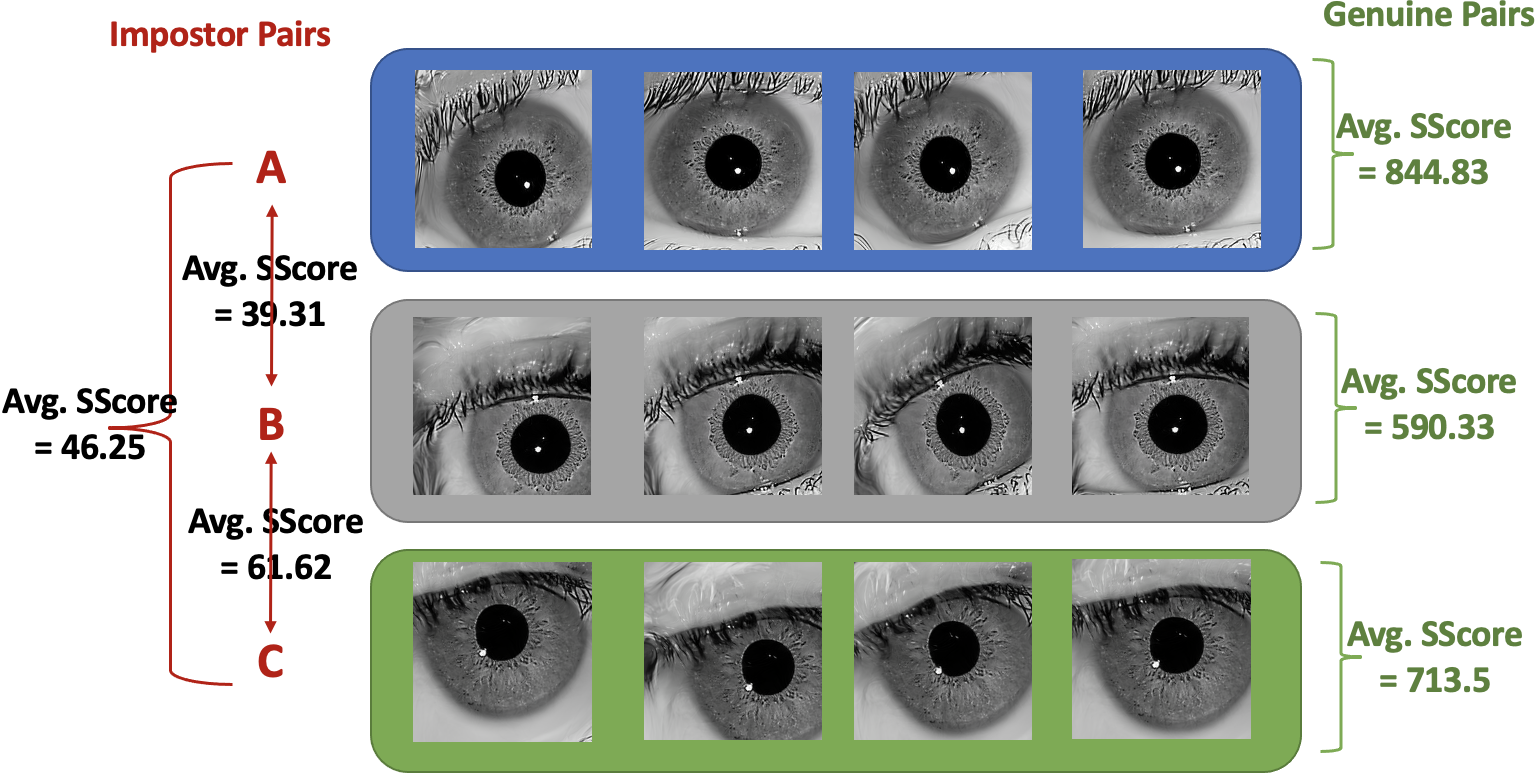}
    \caption{Examples of images generated using iWarpGAN with unique identities and intra-class variations. A total of 20,000 irides corresponding to 2,000 identities were generated for each of the three training datasets. The figure shows the average similarity score (SScore) for both inter and intra class.}
    \vspace{-5mm}
    \label{fig:generated}
\end{figure}

\section{Background}
Generative Adversarial Networks (GANs) \cite{goodfellow2020} are generative models that typically take a random noise vector as input and output a visually realistic synthetic image. A GAN consists of two main components: (1) Generative Network known as Generator (\ensuremath{G}), and (2) Discriminative Network known as Discriminator (\ensuremath{D}) that are in competition with each other. The Generator aims to generate  realistic looking images that can fool the discriminator, while Discriminator (\ensuremath{D}) aims to distinguish between real and synthetic images generated by \ensuremath{G}. In the literature, different methods have been proposed to generate generate good quality biometric images such as face, iris and fingerprint. Some of these methods are discussed below: \\


\noindent
\textbf{Generation using Random Noise:} Kohli et. al. \cite{kohli2017} proposed a GAN-based approach to synthesize cropped iris images using iris Deep Convolution Generative Adversarial Network (iDCGAN). While this method generates good quality cropped iris images of size 64\ensuremath{\times}64, unrealistic distortions and noise were observed when trained to generate high resolution images. In \cite{yadav2019}, Yadav et. al. overcame this issue by utilizing Relativistic Average Standard Generative Network (RaSGAN) that aims to generate good quality high resolution iris images. However, since RaSGAN generates synthetic images from a random noise vector, it is hard to generate irides with intra-class variations. Also, as shown in Figure \ref{fig:Exp1}, the uniqueness of generated images is limited and the network was often observed to repeat certain patterns, restricting the diversity in the generated dataset. Wang et. al. \cite{wang2022} proposed a method for generating iris images that exhibit a wide range of intra- and inter-class variations. Their approach incorporates contrastive learning techniques to effectively disentangle identity-related features, such as iris texture and eye orientation, from condition-variant features, such as pupil size and iris exposure ratio, in the generated images. While their method seems promising but the experiments presented in their paper \cite{wang2022} are not sufficient to comment on quality of iris and uniqueness of the generated images. \\

\noindent    
\textbf{Generation via Image Translation:} Image translation refers to the process of translating an image from one domain to another by learning the mapping between various domains. Therefore, image translation GANs focus on translating a source image to the target domain with the purpose of either changing some style attribute in the source image or adding/mixing different styles together. For example, StyleGAN \cite{karras2019} learns a mapping to different styles in face images (such as hair color, gender, expression, etc.) using a non-linear mapping function that embeds the style code of the target domain into the generated image. Unlike StyleGAN, StarGAN \cite{choi2020} and CIT-GAN \cite{yadav2021} require paired training data to translate a source image to an image with the attributes of the target domain using style code of a reference image. This forces the generator to learn mappings across various domains, making it scalable to multiple domains. However, when trained using real iris images, StarGAN and CIT-GAN were seen to assume the identity of the source image (as shown in Figures \ref{fig:Exp2-Casia-CS.png} and \ref{fig:Exp2-IITD.png}). So, both methods fail to generate irides whose identities are not present in the training dataset.

There are other GAN-based methods in the literature that aim to edit certain portions of the image using warp fields or color transformations. Warp fields have been widely used for editing images such as modifying eye-gaze \cite{ganin2016}, semantically adding objects to an image \cite{zhou2016}, reconstructing facial features \cite{yeh2016}, etc. Dorta et. al \cite{dorta2020} argues that warp fields are more comprehensive than pixel differences that allow more flexibility in terms of partial edits. Geng et. al. \cite{geng2018} proposed WG-GAN that aims to fit a dense warp field to an input source image to translate it according to the target image. This method showed good results at low resolution, but the quality of synthetic data deteriorates at high resolution. Also, as mentioned earlier, the source-target relationship in WG-GAN can restrict the uniqueness of the output image. Dorta et. al. \cite{dorta2020} overcame these issues by proposing WarpGAN that allows partial edits without the dependency on the source-target image pair. The generator takes as input a source image and a target attribute vector and then learns the warp field to make the desired edits in the source image. This method has been proven to make more realistic semantic edits in the input image than StarGAN and CycleGAN \cite{zhu2017}. Further, with the ability of controlled or partial edits, WarpGAN provides the mechanism to generate images with intra-class variations. However, using a real image as input to the generator restricts the number of unique images that can be generated from this network.


\section{Proposed Method}
In this section, we will discuss the proposed method, iWarpGAN, that has the capability to synthesize an iris dataset in such a way that: (1) it contains iris images with unique identities that are not seen during training, (2) generates multiple samples per identity, (3) it is scalable to hundred thousand unique identities, and (4) images are generated in real-time.

Let \ensuremath{x^{d1}_{s1} \in \mathbb{P}} be an input image with identity \ensuremath{d1} and style \ensuremath{s1}, and another input image \ensuremath{x^{d2}_{s2} \in \mathbb{P}} with identity \ensuremath{d2} and style \ensuremath{s2}. Here, \ensuremath{s1} and \ensuremath{s2} denote image with attribute \ensuremath{y}. The attribute vector $y$ is a 12-bit binary vector, where the first 5 bits correspond to a one-hot encoding of angle, the next 5 bits correspond to a one-hot encoding of position shift, and the last 2 bits denote contraction and dilation, respectively. Here, angle and position define eye orientation and the shift of iris center in the given image. The possible angles are \ensuremath{0^o, 10^o, 12^o, 15^o, 18^o} and the possible position shifts are [0,0], [5,5], [10,10], [-10,10], [-10,-10]. For example, an image with angle \ensuremath{10^o}, position shift [0,0] and dilation, the attribute vector \ensuremath{y} will be [0,1,0,0,0,1,0,0,0,0,0,1]. The {\em angle}  value defines the image orientation and {\em position} defines the offset of the iris center from the image center. Given \ensuremath{x^{d1}_{s1}} and \ensuremath{x^{d2}_{s2}}, our aim is to synthesize a new iris image \ensuremath{x^{d3}_{s2}} with identity \ensuremath{d3} different from the training data and possessing the style attribute \ensuremath{s2} from \ensuremath{x^{d2}_{s2}}. To achieve this, as shown in Figure \ref{fig:iWarpGAN}, the framework of iWarpGAN has been divided into five parts: (1) Style Encoder, \ensuremath{E_S}, that encodes style of the input image, (2) Identity Encoder, \ensuremath{E_D}, that learns an encoding to generate an identity different from the input image, (3) Generative Network, \ensuremath{G}, that uses encoding from both \ensuremath{E_D} and \ensuremath{E_S} to generate an image with a unique identity and the given style attribute, (4) Discriminator, \ensuremath{D}, that predicts whether the image is real or synthetic and emits an attribute vector \ensuremath{y^{'}} and (5) Pre-trained Classifier, \ensuremath{C}, that returns the distance score between a real input image and new the identity generated by \ensuremath{G}.

\subsection{Disentangling Identity and Style to Generate New Iris Identities}
Generally, the number of samples available in the training dataset is limited. This restricts the latent space learned by \ensuremath{G} thereby limiting the number of unique identities generated by the trained GAN. Some GANs focus too much on editing or modifying style attributes in the images while generating previously seen identities in the training dataset. This motivated us to divide the problem into two parts: (1) Learning new identities that are different from those in the training dataset, and (2) Editing style attributes for ensuring intra-class variation. Inspired by \cite{zeno2019}, we achieve this by training the proposed GAN using two pathways - Style Transformation Pathway and Identity Transformation Pathway.

\noindent
\textbf{Style Transformation Pathway:}
Similar to StyleGAN, this pathway entirely focuses on learning the transformation of the style. Therefore, this sub-path aims to train the networks \ensuremath{E_S, D} and \ensuremath{G}, while keeping the networks \ensuremath{E_D} and \ensuremath{C} fixed. Input to the generator \ensuremath{G} is the concatenated latent vector \ensuremath{d} and \ensuremath{s} to generate an iris image with style attribute \ensuremath{y}. \ensuremath{G} tries to challenge \ensuremath{G} by maximizing,
\vspace{-2mm}

{\small
\begin{equation}\label{eq4}
    \ensuremath{\mathcal{L}_{G-Sty} = \mathbb{E}_{x^{di}_{si}, x^{dj}_{sj} \sim \mathbb{P}_{real}}[D(G(E_D(x^{di}_{si}), E_S(x^{dj}_{sj}, y)))]}
\end{equation}
}

Here, \ensuremath{\bar{x} = (G(E_D(x^{di}_{si}), E_S(x^{dj}_{sj}, y)))} is the image generated by \ensuremath{G}. At the same time, \ensuremath{D} competes with \ensuremath{G} by minimizing,
\vspace{-1mm}

{\small
\begin{equation}\label{eq5}
\begin{split}
    \ensuremath{\mathcal{L}_{D-Sty} = \mathbb{E}_{x^{di}_{si}, x^{dj}_{sj} \sim \mathbb{P}_{real}} [D(G(E_D(x^{di}_{si}), E_S(x^{dj}_{sj}, y)))] \\
    - \mathbb{E}_{x}[D(x)]}
\end{split}
\end{equation}
}

In order to enforce that an iris image is generated with style attributes \ensuremath{y}, the following loss function is utilized:
\vspace{-1mm}

{\small
\begin{equation}\label{eq6}
    \ensuremath{\mathcal{L}_{Sty-Recon} = || E_S(\bar{x}) - E_S(x^{dj}_{sj}) ||_2^2}
\end{equation}
}
\vspace{-1mm}

\begin{figure*}
\centering
\subfloat[][CASIA-Iris-Thousand dataset v/s synthetically generated images from different GANs]{%
  \includegraphics[width=0.95\textwidth]{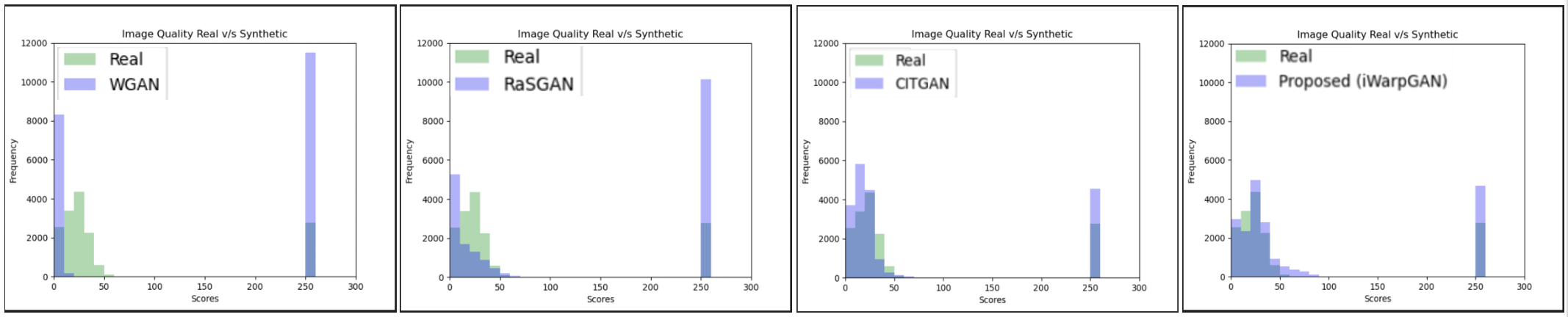}%
} \\
\subfloat[][CASIA-CSIR dataset v/s synthetically generated images from different GANs]{%
  \includegraphics[width=0.95\textwidth]{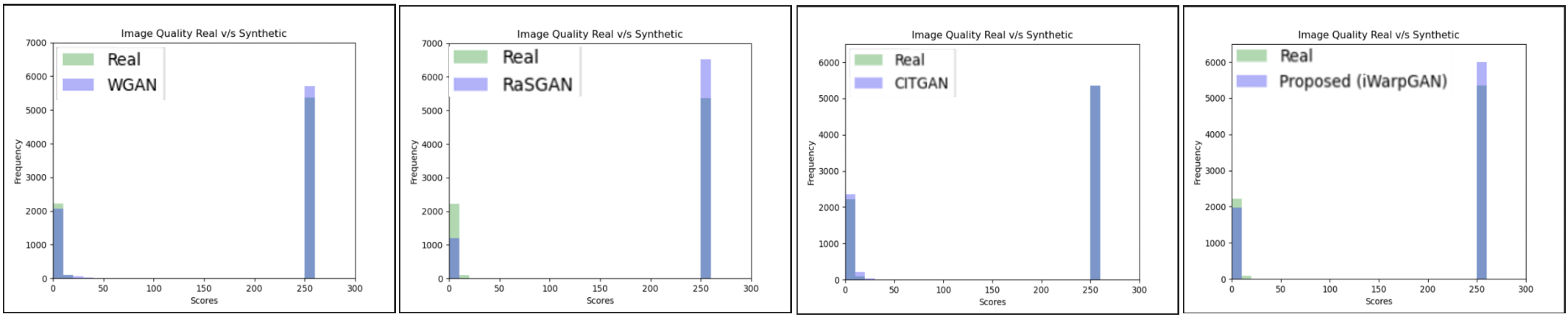}%
} \\
\subfloat[][IITD-iris dataset v/s synthetically generated images from different GANs]{%
  \includegraphics[width=0.95\textwidth]{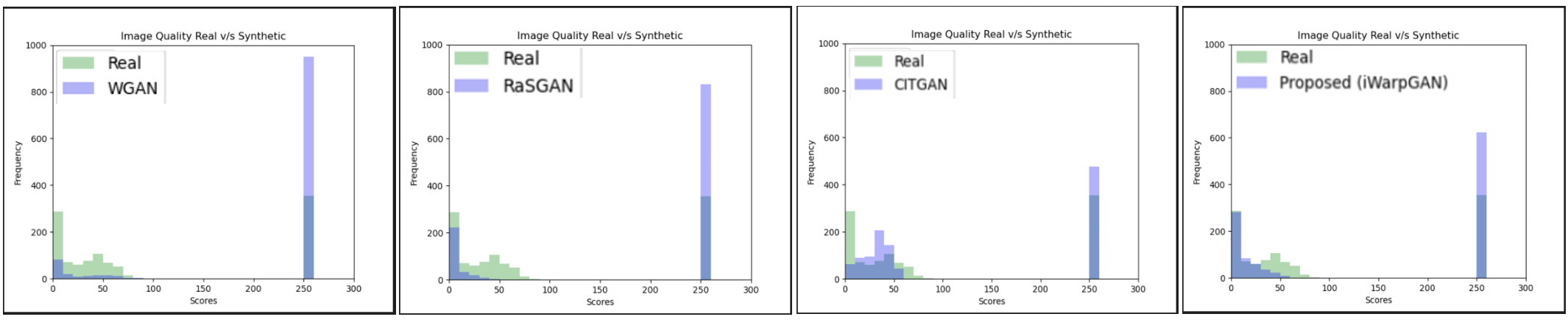}%
} \\
\caption{Histograms showing the quality scores of real iris images from three different datasets and the synthetically generated iris images. The quality scores were are generated using ISO/IEC 29794-6 Standard Quality Metrics\cite{iso-quality} in the score range of [0-100]. Higher the score, better the quality. Iris images that failed to be processed by this method are given the score of 255.}
\vspace{-5mm}
\label{fig:Exp1}
\end{figure*}

\noindent
\textbf{Identity Transformation Pathway:}
This pathway focuses on learning identities in latent space that are different from the training dataset. Therefore, this sub-path aims to train the networks \ensuremath{E_D, D} and \ensuremath{G}, while keeping the networks \ensuremath{E_S} and \ensuremath{C} fixed. Therefore, 

\vspace{-4mm}
{\small
\begin{equation}\label{eq7}
    \ensuremath{\mathcal{L}_{G-ID} = \mathbb{E}_{x^{di}_{si}, x^{dj}_{sj} \sim \mathbb{P}_{real}}[D(G(E_D(x^{di}_{si}), E_S(x^{dj}_{sj}, y)))]}
\end{equation}
}

{\small
\begin{equation}\label{eq8}
\begin{split}
    \ensuremath{\mathcal{L}_{D-ID} = \mathbb{E}_{x^{di}_{si}, x^{dj}_{sj} \sim \mathbb{P}_{real}} [D(G(E_D(x^{di}_{si}), E_S(x^{dj}_{sj}, y)))] \\
    - \mathbb{E}_{x}[D(x)]}
\end{split}
\end{equation}
}

Here, the goal is to learn encodings that represent identities different from those in the training dataset. For this, \ensuremath{E_D} is divided into two parts (as shown in Figure \ref{fig:iWarpGAN}) - Encoder \ensuremath{E} that extracts the encoding from given input image and Warping Network \ensuremath {W} that aims to learn \ensuremath{M} warping functions (\ensuremath{f^1,.....,f^M}) to discover \ensuremath{M} non-linear paths in the latent space of \ensuremath{G}. The gradient of these can be utilized to define the direction at each latent code \ensuremath{z} \cite{tzelepis2021} such that new \ensuremath{\bar{z}} represents encoding of an identity different from the input image. In order to achieve this, the encoder \ensuremath{E_D} is broken down to two parts - an encoder \ensuremath{E} that extracts the latent code of the given input image and passes it on to the warping network \ensuremath{W}.

For a vector space \ensuremath{\mathbb{R}^d}, the function \ensuremath{f:\mathbb{R}} is defined as,

\begin{equation}\label{eq9}
    \ensuremath{ f(z) = \sum_{k=1}^{K} b_i exp (-u_i  || z - v_i ||^2) } 
\end{equation}

Here, \ensuremath{v_i \in \mathbb{R}^d} represents the center, \ensuremath{b_i \in \mathbb{R}} represents weight and \ensuremath{u_i \in \mathbb{R}} represents scale of \ensuremath{i^{th}} RBF. This function for warping is differentiable and for a specific value of \ensuremath{z}, the direction from \ensuremath{\Delta f} can be used to define a curve in \ensuremath{\mathbb{R}^d} by shifting \ensuremath{z} as \cite{tzelepis2021}:

\begin{equation}\label{eq10}
    \ensuremath{ \delta z = \epsilon \frac{\Delta f(z)}{||\Delta f(z)||} } 
\end{equation}

Here, \ensuremath{\epsilon} is the shift magnitude that determines the shift from \ensuremath{z} to \ensuremath{\bar{z}} using above equation. The Warping Network, \ensuremath{W}, contains two components: warper and reconstructor \ensuremath{R}. The warper can be parameterized using the triplet \ensuremath{(V^m, B^m, U^m)} denoting the center, weight and parameters. Here, \ensuremath{m = 1,2,....M} and each triplet help warping the latent space in \ensuremath{\mathbb{R}^d}. Also, the reconstructor is utilized to estimate the support set and magnitude shift that led to the transformation at hand. Therefore, the objective function for the Warping Network can be defined as,

 \begin{equation}\label{eq11}
    \ensuremath{ \min\limits_{V,B,U,R} \mathbb{E}_{z,\epsilon} [ \mathcal{L}_{W-Reg} (\epsilon, \bar{\epsilon})] 
    } 
\end{equation}

Here, \ensuremath{\mathcal{L}_{W-Reg}} refers to regression loss. To further emphasize the uniqueness of identity learned by \ensuremath{G} in latent space, we maximize,

 \begin{equation}\label{eq12}
    \ensuremath{ \mathcal{L}_{Ident-Recon} = || E_D(\bar{x}) - E(x^{di}_{si}) ||_2^2
    } 
\end{equation}

\begin{equation}\label{eq13}
    \ensuremath{ \mathcal{L}_{Ident-Cls} = || Feat(\bar{x}) - Feat(x^{di}_{si}) ||_2^2
    } 
\end{equation}

Here, \ensuremath{Feat(x)} are the features extracted by the trained iris classifier (i.e., matcher) \ensuremath{C}.

By employing distinct pathways for style and identity, the proposed method enables the manipulation of identity features to generate synthetic images with distinct identities that diverge from the training dataset. Additionally, this methodology allows for the generation of images with varied styles for each identity. This is achieved by keeping the input image to the identity pathway constant and varying the input image  to the style pathway to enforce that the generated images have the same identity \ensuremath{d} but different styles (i.e., intra-class variation) \ensuremath{s_1, s_2, ....,s_n}.


\section{Datasets}
In this work, we utilized three publicly available iris datasets for conducting experiments and performing our analysis:

\noindent
\textbf{D1: CASIA-Iris-Thousand}
This dataset \cite{casia1000} released by the Chinese Academy of Sciences Institute of Automation has been widely used to study distinctiveness of iris features and to develop state-of-the-art iris recognition methods. It contains 20,000 irides from 1,000 subjects (2,000 unique identities with left and right eye) captured using an iris scanner with a resolution of 640\ensuremath{\times}480. The dataset is divided into train and test sets using a 70-30 split based on unique identities, i.e., 1,400 identities in the training set and 600 in the test set.

\noindent
\textbf{D2: CASIA Cross Sensor Iris Dataset (CSIR)}
For this work, we had access to only the train set of the CASIA-CSIR dataset \cite{xiao2013} released by the Chinese Academy of Sciences Institute of Automation. This dataset consists of 7,964 iris images from 100 subjects (200 unique identities with left and right eye), which is divided into train and test sets using a 70-30 split on unique identities for training and testing deep learning based iris recognition methods, i.e., training set contains 5,411 images and test set contains 2,553 images.

\noindent
\textbf{D3: IITD-iris}
This dataset  \cite{iitd-iris} was released by the Indian Institute of Technology, Delhi, and was acquired in an indoor environment. It contains 1,120 iris images from 224 subjects captured using JIRIS, JPC1000 and digital CMOS cameras with a resolution of 320\ensuremath{\times}240. This dataset is divided into train and test sets using 70-30 split based on unique identities, i.e., images from 314 identities in the training set and images from 134 identities in the testing set. \\

\noindent
\textbf{Training Data for Proposed Method}
The proposed method is trained using cropped iris images of size 256\ensuremath{\times}256, where the style of each image is represented using the attribute vector \ensuremath{y}. Current datasets do not contain balanced number of iris images across these attributes. Therefore, variations such as angle and position is added via image transformations on randomly selected images from the dataset. In order to achieve this, first iris coordinates are first obtained using the VeriEye iris matcher, images are then translated to different angles and positions with respect to these centers, and cropped iris image of size 256\ensuremath{\times}256 extracted. This helps create a training dataset with balanced samples across different attributes. Since the proposed method uses an image translation GAN, during image synthesis two images \ensuremath{x^{d1}_{s1}}, \ensuremath{x^{d2}_{s2}} and an attribute vector \ensuremath{y} of image \ensuremath{x^{d2}_{s2}} are used as input to synthesize a new iris image \ensuremath{x^{d3}_{s2}} with identity \ensuremath{d3} which is different from the training data and possesses the style attribute \ensuremath{s2} of \ensuremath{x^{d2}_{s2}}.


\begin{figure}
    \centering
    \captionsetup{justification=centering}
    \includegraphics[width=0.45\textwidth]{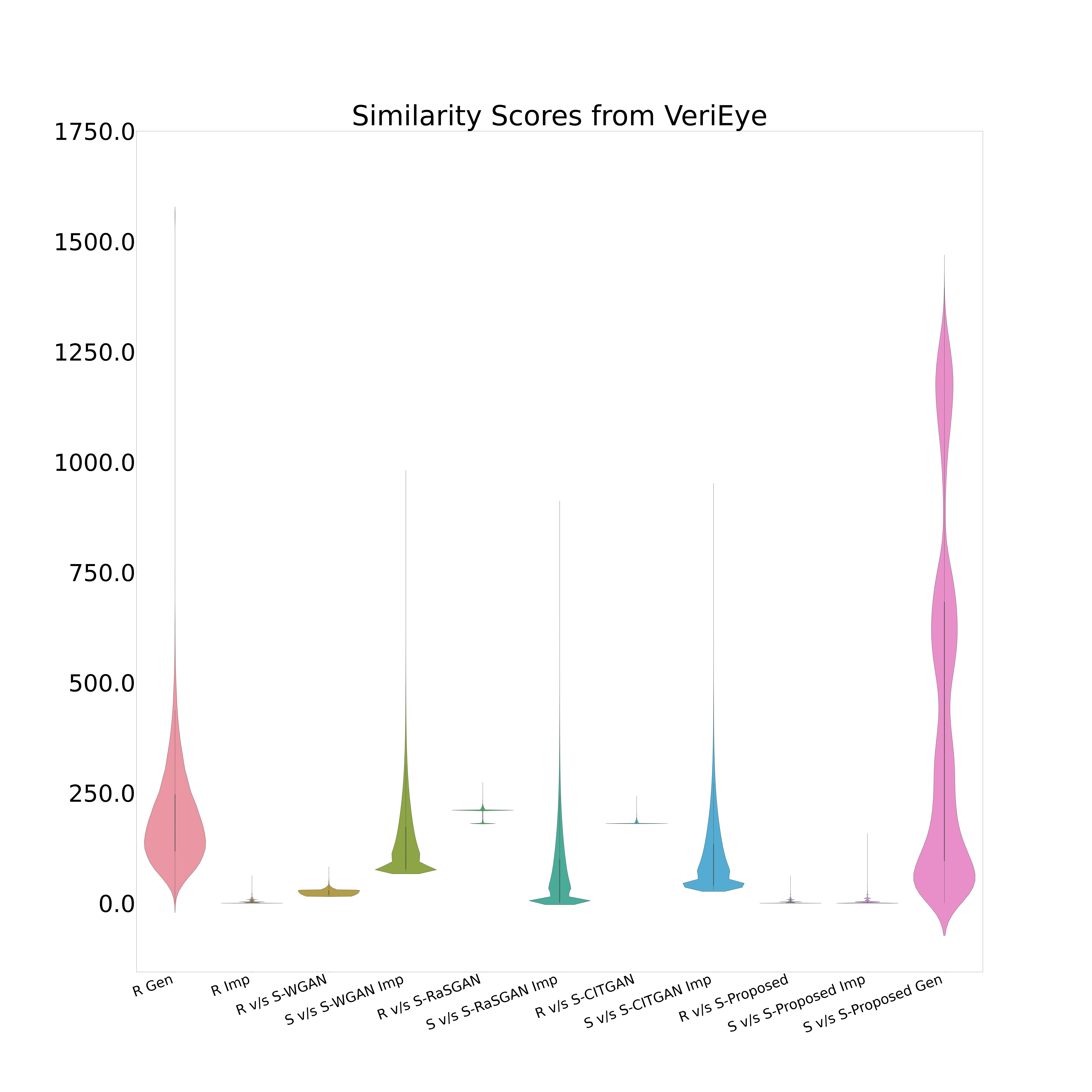}
    \caption{This figure shows the uniqueness of iris images generated using iWarpGAN when the GANs are trained using CASIA-Iris-Thousand dataset. The y-axis represents the similarity scores obtained using VeriEye. Here, R=Real, S=Synthetic, Gen=Genuine and Imp=Impostor.}
    \vspace{-5mm}
    \label{fig:Exp2-Casia1000.png}
\end{figure}

\begin{figure}
    \centering
    \captionsetup{justification=centering}
    \includegraphics[width=0.45\textwidth]{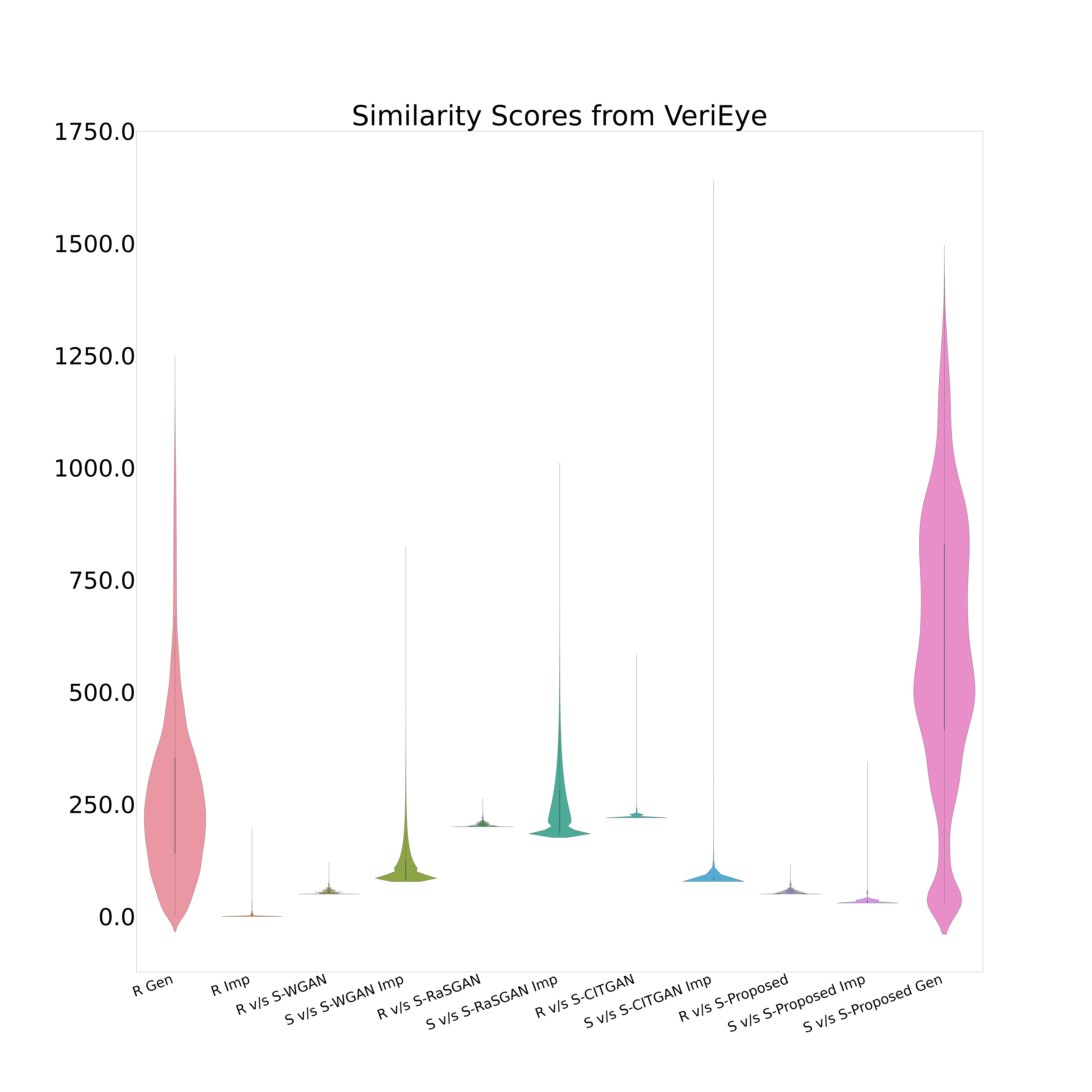}
    \caption{This figure shows the uniqueness of iris images generated using iWarpGAN when the GANs are trained using CASIA-CS iris dataset. The y-axis represents the similarity scores obtained using VeriEye. Here, R=Real, S=Synthetic, Gen=Genuine and Imp=Impostor.}
    \vspace{-5mm}
    \label{fig:Exp2-Casia-CS.png}
\end{figure}

\begin{figure}
    \centering
    \captionsetup{justification=centering}
    \includegraphics[width=0.45\textwidth]{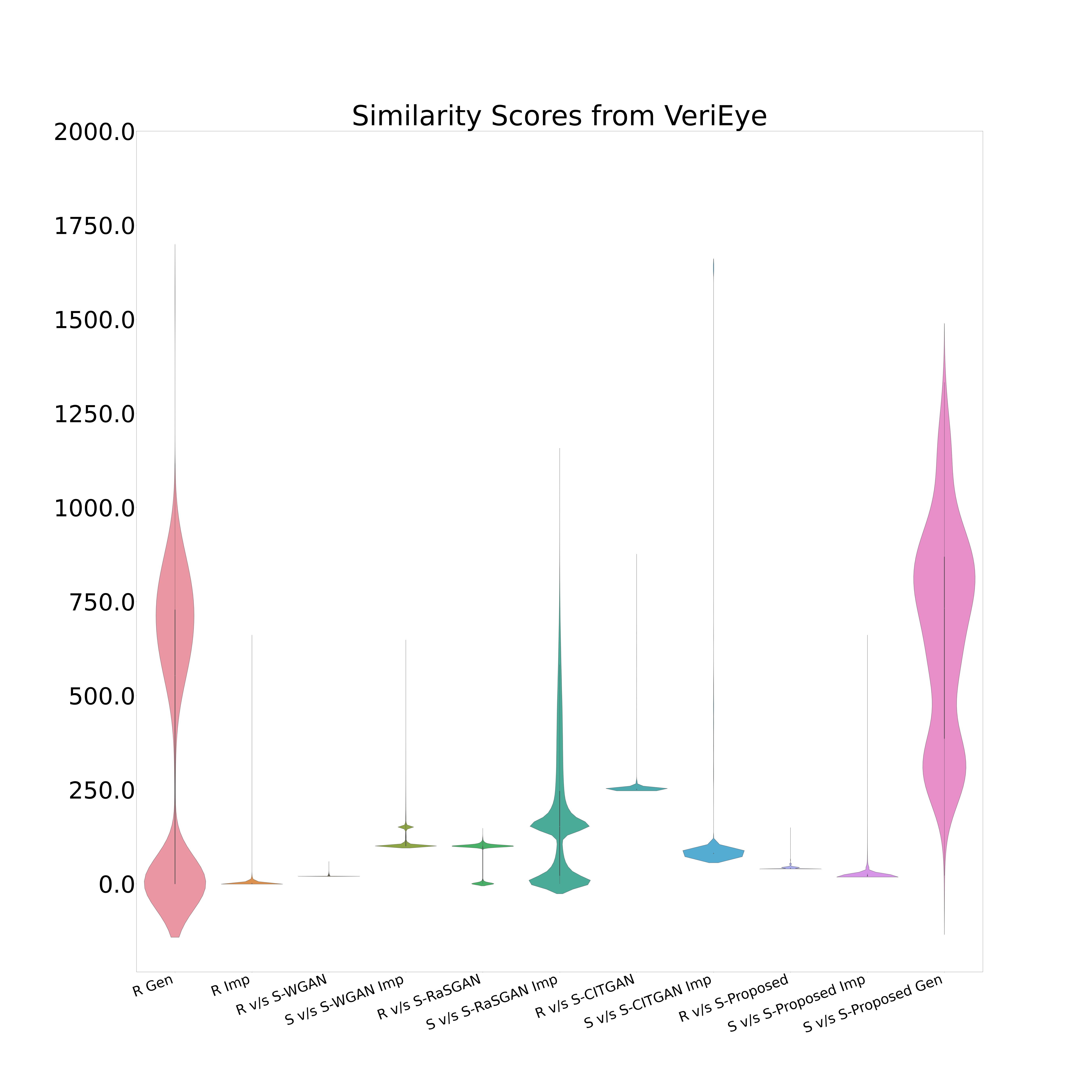}
    \caption{This figure shows the uniqueness of iris images generated using iWarpGAN when the GANs are trained using IITD iris dataset. The y-axis represents the similarity scores obtained using VeriEye. Here, R=Real, S=Synthetic, Gen=Genuine and Imp=Impostor.}
    \vspace{-5mm}
    \label{fig:Exp2-IITD.png}
\end{figure}

\begin{figure*}
    \centering
    \captionsetup{justification=centering}
    \includegraphics[width=0.85\textwidth]{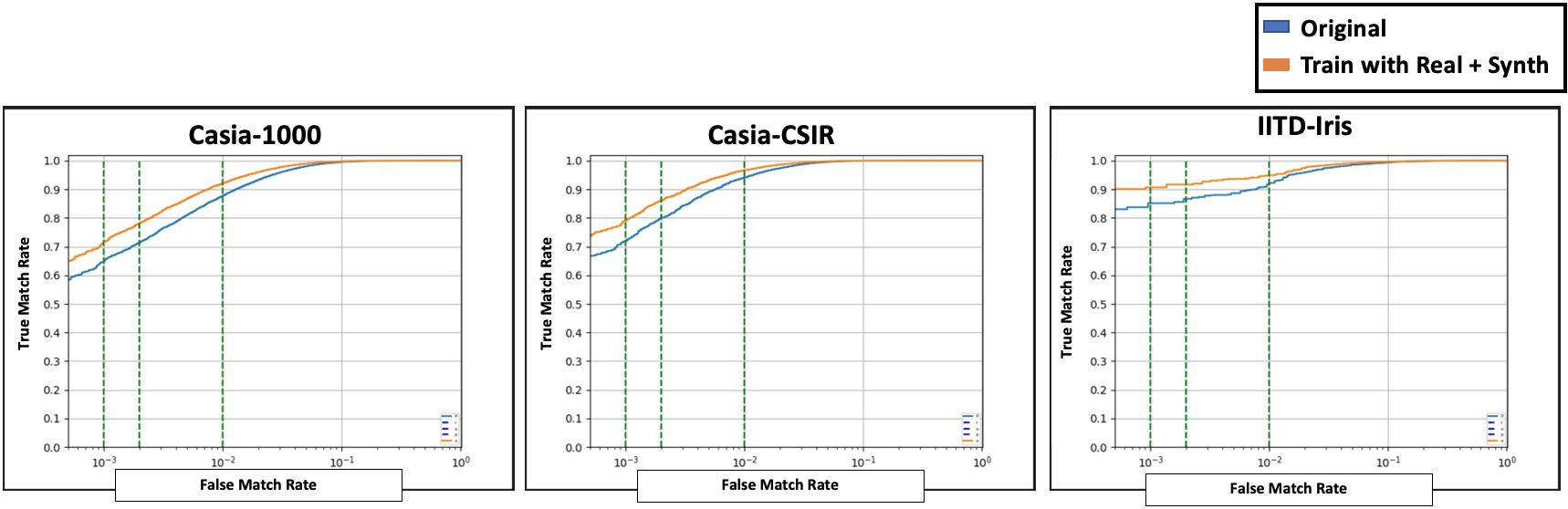}
    \caption{This figure shows the performance of Resnet-101 in the cross-dataset evaluation scenario. (a) Trained using train set of CASIA-CSIR \& IIT-Delhi datasets and tested using test set of CASIA-Iris-Thousand. (b) Trained using CASIA-Iris-Thousand \& IIT-Delhi datasets and tested using test set of CASIA-CSIR dataset. (c) Trained using CASIA-Iris-Thousand \& CASIA-CSIR datasets and tested using test set of IIT-Delhi iris dataset.}
    \vspace{-5mm}
    \label{fig:Resnet101-Exp3}
\end{figure*}

\begin{figure*}
    \centering
    \captionsetup{justification=centering}
    \includegraphics[width=0.85\textwidth]{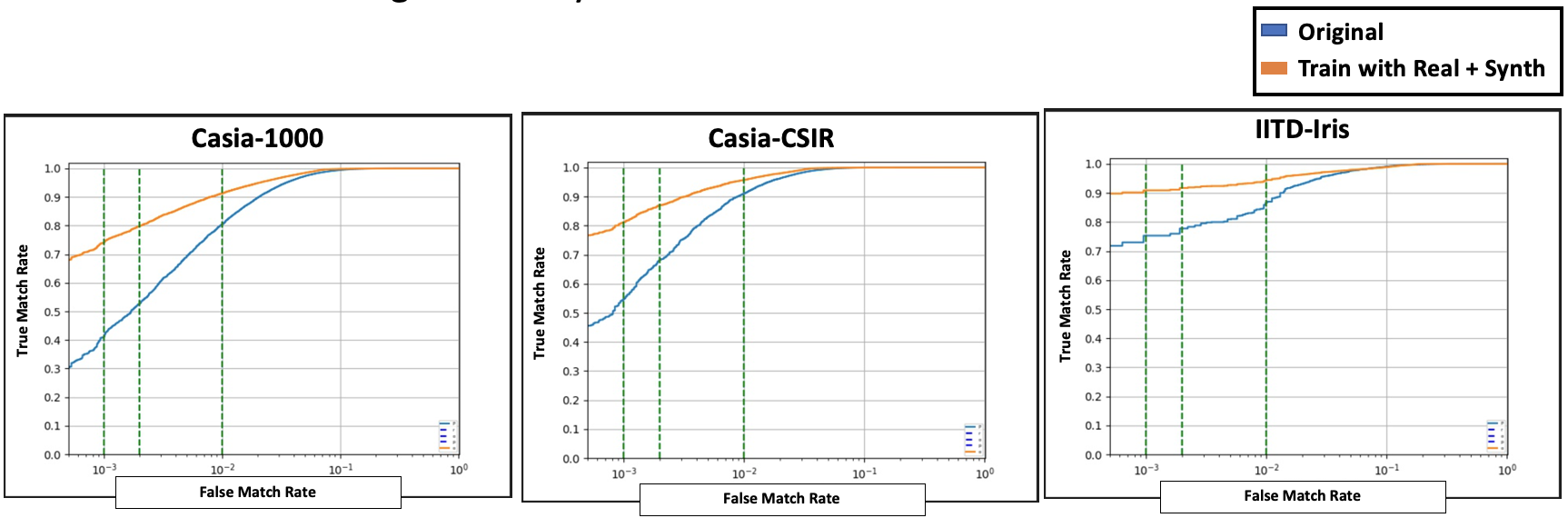}
    \caption{This figure shows the performance of EfficientNet in the cross-dataset evaluation scenario. (a) Trained using train set of CASIA-CSIR \& IIT-Delhi datasets and tested using test set of CASIA-Iris-Thousand. (b) Trained using CASIA-Iris-Thousand \& IIT-Delhi datasets and tested using test set of CASIA-CSIR dataset. (c) Trained using CASIA-Iris-Thousand \& CASIA-CSIR datasets and tested using test set of IIT-Delhi iris dataset.}
    \vspace{-5mm}
    \label{fig:EfficientNet-Exp3}
\end{figure*}

\section{Experiments \& Results}
\label{sec:experiments}
In this section, we discuss different experiments utilized to study and analyze the performance of the proposed method. First, three sets of 20,000 number of iris images corresponding to 2,000 identities are generated. The three sets correspond to three different training datasets, D1, D2 and D3. For some of the experiments below, a subset of the generated images were used in order to be commensurate with the corresponding real dataset.


\subsection{Experiment-1: Quality of Generated Images}

\noindent
\textbf{ISO/IEC 29794-6 Standard Quality Metrics}
The quality of generated images is compared with the real images using ISO/IEC 29794-6 Standard Quality Metrics \cite{iso-quality}. We also evaluated the quality of images generated by other techniques, viz., WGAN \cite{arjovsky2017}, RaSGAN \cite{yadav2019} and CITGAN \cite{yadav2021} and compared them with the images generated using iWarpGAN. The ISO metric evaluates the quality of an iris image using factors such as usable iris area, iris-sclera contrast, sharpness, iris-pupil contrast, pupil circularity, etc. to generate an overall quality score. The quality score ranges from [0-100] with 0 representing poor quality and 100 representing the highest quality. The images that cannot be processed by this method (either due to extremely poor quality or error during segmentation) are given a score of 255.

As shown in Figure \ref{fig:Exp1}, the quality scores of iris images generated by iWarpGAN and CITGAN are comparable with real irides. On the other hand, WGAN and RaSGAN have many images with a score of 255 due to the poor image quality. Also, when comparing the images in the three datasets, it can be seen that CASIA-CSIR dataset has more images with a score of 255 than IITD-iris and CASIA-Iris-Thousand dataset.
\vspace{2mm}

\noindent
\textbf{VeriEye Rejection Rate}
To further emphasize the superiority of the proposed method in generating good quality iris images, we compare the rate of rejection of the generated images  by a commercial iris matcher known as VeriEye. We compare the rejection rate for images generated by iWarpGAN with the real images as well as  those generated by WGAN, RaSGAN and CITGAN:
\vspace{2mm}

\noindent
(a) IITD-Iris-Dataset: This dataset contains a total of 1,120 iris images out of which 0.18\% images are rejected by VeriEye. For comparison, we generated 1,120 iris images each using iWarpGAN, WGAN, RaSGAN and CITGAN. For the generated images, the rejection rate is as high as 9.73\% and 4.55\% for WGAN and RaSGAN, respectively. However, the rejection rate for CITGAN and iWarpGAN is 2.85\% and 0.73\%, respectively.
\vspace{2mm}

\noindent
(b) CASIA-CS Iris Dataset: This dataset contains a total of 7,964 iris images out of which 2.81\% images are rejected by VeriEye. For comparison, we generated 7,964 iris images each using iWarpGAN, WGAN, RaSGAN and CITGAN. For the generated images, the rejection rate is as high as 4.17\% and 2.06\% for WGAN and RaSGAN, respectively. However, the rejection rate for CITGAN and iWarpGAN is 2.71\% and 2.74\%, respectively.
\vspace{2mm}

\noindent
(c) CASIA-Iris-Thousand Dataset: This dataset contains a total of 20,000 iris images out of which 0.06\% images are rejected by VeriEye. For comparison, we generated 20,000 iris images each using iWarpGAN, WGAN, RaSGAN and CITGAN. For the generated images, the rejection rate is as high as 0.615\% and 0.34\% for WGAN and RaSGAN, respectively. However, the rejection rate for CITGAN and iWarpGAN is 0.24\% and 0.18\%, respectively.

\subsection{Experiment-2: Uniqueness of Generated Images}
This experiment analyzes the uniqueness of the synthetically generated images, i.e., we evaluate whether iWarpGAN is capable of generating unique identities with intra-class variations.
\vspace{2mm}

\noindent\textbf{Experiment-2A:}
Experiment-2A focuses on studying the uniqueness in the synthetic iris dataset generated using different GAN methods with respect to training samples. For this, we studied the genuine and impostor distribution of real iris images used to train GAN methods and compared it with the distribution of synthetically generated iris images. We utilized VeriEye matcher in this experiment to evaluate the similarity score between a pair of iris image. The score ranges from [0, 1557] where a higher score denotes a better match. 
\vspace{2mm}

\noindent\textbf{Experiment-2B:}
Experiment-2B focuses on studying the uniqueness and intra-class variations within the generated iris dataset. For this, we studied the genuine and impostor distributions of the generated iris images and compare it with the distribution of real iris datasets. As mentioned earlier, this study is done for various unique generated identities to study both uniqueness and scalability. We utilized VeriEye matcher in this experiment to evaluate the similarity score between a pair of iris images.
\vspace{2mm}

\noindent \textbf{Analysis}\\
As shown in the Figures \ref{fig:Exp2-Casia1000.png}, \ref{fig:Exp2-Casia-CS.png} and \ref{fig:Exp2-IITD.png}, unlike other GAN methods, the iris images generated by iWarpGAN do {\em not} share high similarity with the real iris images used in training. This shows that iWarpGAN is capable of generating irides with identities that are different from the training dataset. Further, looking at the impostor distribution of synthetically generated images, which overlaps with the impostor distribution of real iris images, we can conclude that the generated identities are different from each other. Note that low similarity scores in WGAN for real v/s synthetic and synthetic v/s synthetic distributions are due to poor quality iris images generated by WGAN.

\subsection{Experiment-3: Utility of Synthetic Images}
In this experiment, we analyze the performance of deep learning algorithms trained and tested for iris recognition using a triplet training method, and compare it with the performance when these algorithms are trained using real and synthetically generated iris images.
\vspace{2mm}

\noindent \textbf{Experiment-3A: Baseline Analysis}
\\
This is a baseline experiment where EfficientNet \cite{hsiao2021} and Resnet-101 \cite{minaee2019} are trained with the training set of CASIA-Iris-Thousand, CASIA-CSIR and IITD-iris datasets using the triplet training method. The trained networks are tested for iris recognition on the test set of the above mentioned datasets (as mentioned in Section IV).
\vspace{2mm}

\noindent \textbf{Experiment-3B: Cross-Dataset Analysis}
\\
In this experiment, we analyze the benefits of synthetically generated iris datasets in improving the performance of deep learning based iris recognition methods. EfficientNet and Resnet-101 are trained using the training set of CASIA-Iris-Thousand, CASIA-CSIR and IITD-iris datasets, as well as the synthetically generated iris dataset from the iWarpGAN. 
\vspace{2mm}

\noindent \textbf{Analysis:}
As shown in Figures \ref{fig:Resnet101-Exp3} and \ref{fig:EfficientNet-Exp3}, the performance of the deep learning based iris recognition system improves when trained with more data, i.e., when combining real and synthetically generated iris images from iWarpGAN. While there is a slight improvement in the performance of ResNet-101, a significant improvement in the performance is seen for EfficientNet.


\section{Summary \& Future Work}
The results in Section \ref{sec:experiments} show that unlike current GANs, the proposed method is capable of generating good quality iris images with identities that are different from the training dataset. Also, the generated identities are unique with respect to each other with some variations. We also showed the usefulness of the generated dataset in improving the performance of deep learning-based iris recognition methods by providing additional synthetic training data with numerous unique identities. The proposed method is based on image transformation, i.e., the network needs an input and reference image to transform the identity and the style and produce an output image. This can limit the feature space explored by iWarpGAN. For future work, we would like to extensively study the capacity of the proposed method in terms of number of unique identities it can generate and further explore how to make the proposed method more generalizable so that the new identities learnt by iWarpGAN is not limited by the training set. \\ 

\noindent
{\small This research is based upon work supported by by NSF CITeR funding. The views and conclusions contained herein are those of the authors and should not be interpreted as necessarily representing the official policies, either expressed or implied by NSF CITeR.}

\balance
{\small
\bibliographystyle{ieee}
\bibliography{egbib}
}

\end{document}